\documentclass{article}

\usepackage[preprint]{nips_2018}

\usepackage{times}
\usepackage{fixme}
\usepackage{graphbox} %loads graphicx package
\usepackage{subfigure}
\usepackage{amsmath,amssymb,xspace}
\usepackage{natbib}
\usepackage{amsthm}
\usepackage{algorithm}
\usepackage{algorithmic}
\usepackage[]{tikz}
\usepackage[export]{adjustbox}
\usepackage{array}
\usepackage{smartdiagram}
\usetikzlibrary{calc,positioning,shapes.geometric,shadows.blur,arrows,decorations.pathreplacing}
\usepackage[colorlinks=true]{hyperref}
\usepackage[utf8]{inputenc}

\title{ABC-GAN: Easy High-Dimensional Likelihood-Free Inference}
\author{Vinay Jethava, Devdatt Dubhashi}

\begin{document}

\maketitle
\newcommand{\bX}{{\bf X}}
\newcommand{\bY}{{\bf Y}}
\newcommand{\bx}{{\bf x}}
\newcommand{\by}{{\bf y}}
\newcommand{\bz}{{\bf z}}

\newcommand{\PA}{\emph{improve\_approx}\xspace}
\newcommand{\PB}{\emph{improve\_accept}\xspace}
\begin{abstract}
We introduce a framework using Generative Adversarial
Networks (GANs) for likelihood--free inference (LFI) and Approximate Bayesian
Computation (ABC) where we replace the black-box simulator model with an approximator network and generate a rich set of summary features in a data driven fashion.  On benchmark data sets, our approach improves on others with respect to scalability, ability to handle high dimensional data and complex probability distributions.
\end{abstract}

%% Please note - this section has been moved to intro.tex
\section{Introduction} 
%The likelihood function ${\cal L}(\theta) = P({\bf X} \mid \theta)$ plays a central role in statistics and machine learning. 
% However, in several problems, the likelihood function is not available or expensive to compute and Approximate Bayesian Computation~(ABC)~\citep{sisson2018overview} is a popular technique that is used to perform Likelihood-free Inference (LFI).

Approximate Bayesian Computation~(ABC) is a likelihood-free inference method that learns the  parameter $\theta$ by generating simulated data $\bY_{\theta}$ and
accepting proposals (for the parameter $\theta$) when the simulated
data resembles the true data $\bX$~\citep[see][for a recent overview]{lintusaari2017fundamentals}. In most high-dimensional settings, a summary statistic $T$ is chosen to represent the data so that the
distance $d(\bX, \bY_{\theta})$ can be computed in terms of the
summary statistic $d_{T}(t_{\bX}, t_{\bY})$. 
Rejection ABC~\citep{pritchard1999population} often suffers from low acceptance rates and several methods have been developed to address this problem. Broadly, these can be categorized as Markov Chain Monte Carlo methods~\citep{marjoram2003markov,beaumont2009adaptive,meeds2015hamiltonian,moreno2016automatic}, sequential Monte Carlo~\citep{sisson2007sequential}, and more recently, classifier-based approaches based on Bayesian optimization (BOLFI)~\citep{gutmann2016bayesian}.

In a series of recent papers, \citet{gutmann2014statistical,gutmann2017likelihood,gutmann2016bayesian} have suggested two novel ideas: $(1)$ treating the problem of discriminating distributions as a classification problem  between $\bX$ and $\bY_{\theta}$; and $(2)$
regression of the parameter $\theta$ on the distance-measure $d_{T}(\cdot, \cdot)$ using Gaussian Processes in order to identify the suitable regions of the parameter space having higher acceptance ratios. 

This line of work is closely related to Generative Adversarial Networks~(GANs)~\citep{goodfellow2014generative} with different flavors of GANs~\citep{goodfellow2014generative,nowozin2016f,arjovsky2017wasserstein,dziugaite2015training} minimizing alternative divergences (or ratio losses) between the observed data $\bX$ and simulated data $\bY_\theta$~\citep[see][for an excellent exposition]{mohamed2016learning}. 

In this paper, we develop the connection between GANs~\citep{goodfellow2016nips} and \cite{gutmann2017likelihood} and present a new differentiable architecture inspired by GANs for likelihood-free inference~(Figure~1).  In doing so, we make the following contributions: 
\begin{itemize} 
    \item We present a method for adapting  black box simulator-based model with an ``approximator", a differentiable neural network module~\citep[similar to function approximation, see e.g.,][and references therein]{liang2016deep}. 
    \item Our method provides automatic generation of summary statistics as well as the choice of different distance functions (for comparing the summary statistics) with clear relation to likelihood ratio tests in GANs.
    \item  We adapt one of the key ideas in \cite{gutmann2017likelihood}, namely, gradient-descent based search to quickly narrow down to the acceptance region of the parameter space, to the framework of GANs.  
\item We perform experiments on real-world problems (beyond Lotka-Volterra model) previously studied in the ABC literature - showing benefits as well as cases where the differentiable neural network architecture might \emph{not} be the best solution~(Section~\ref{sec:experiments}). 
\end{itemize}
\section{Related Work}
\label{sec:related-work}
%ABC is a rich field with theoretical maturity~\citep[see e.g.][and other chapters]{sisson2018overview} and applications in several domains. This section presents three lines of work closely related to this paper. 

\paragraph{Summary statistics and distance function} 
The choice of summary statistic is known to be critical to the 
performance of the ABC scheme~\citep{blum2013comparative,prangle2015summary}. Several methods have explored automatic generation of summary statistics, e.g., projection using regression~\citep{fearnhead2012constructing}, random forests~\citep{marin2016abc}, etc.
More recently, \citet{prangle2017adapting} have explored alternative distance functions for comparison of summary statistics. 
Within the classification scheme of \citep{prangle2015summary,prangle2017adapting}, one of our contributions is automatic computation of non-linear projection-based summary statistics \emph{and} a moment matching distance function~(respectively, MMD~\citep{dziugaite2015training} or Wasserstein distance~\citep{arjovsky2017wasserstein}).

% \paragraph{Classical ABC methods}
% Several approaches have addressed the poor acceptance rates of rejection-sampling ABC. Broadly, these can be categorized as Markov Chain Monte Carlo methods~\citep{marjoram2003markov,beaumont2009adaptive,moreno2016automatic}, sequential Monte Carlo~\citep{sisson2007sequential}, and more recently, classifier-based approaches based on Bayesian optimization (BOLFI)~\citep{gutmann2016bayesian}.

% \paragraph{Function approximation}
% Considerable work has been done on function approximation using neural networks~\citep[see e.g.,][and references therein]{liang2016deep}. The ``approximator" unit in our work can be seen as a function approximator (though not approximating the black-box simulator output directly, rather, the automated summary statistics generated based on the simulator output). 

\paragraph{LFI using neural networks}
Several authors have recently proposed alternative approaches for likelihood-free inference~\citep{DBLP:journals/corr/MeedsW15,cranmer2015approximating,papamakarios2016fast,tran2017hierarchical}. 
In particular, \citet{papamakarios2016fast} inverted the ABC problem by sampling the parameter $\theta$ from mixture of Gaussians $q_\phi(\theta| \bx)$ (parametrized by neural network model $\phi$). 
More recently, \citet{tran2017hierarchical} presented an elegant variational approach for likelihood-free inference
%minimizing the KL-divergence between true probability  $p(\bx, \bz, \beta ) = p(\beta)\prod p(\bx_n| \bz_n, \beta) p(\bz_n | \beta )$ and simpler distribution $q(\bx, \bz, \beta)$ 
under the restrictions that (1) the conditional density $p(\bx_n | \bz_n , \theta)$ is indirectly specified as the hierarchical implicit model (similar to the ``approximator" in this work) which given input noise $\epsilon_n\sim s(\cdot)$ outputs sample $\bx_n$ given latent variable $\bz_n$ and parameter $\theta$, i.e.,  $\bx \sim g(\epsilon | \bz, \theta), \quad \epsilon_n \sim s(\cdot)$; and (2) one can sample from the target distribution $q(\bz_n | \bx_n, \theta)$.

% This allows use of ratio test (similar to GANs) and alternate optimization. 
% Following \citet{papamakarios2016fast}, the authors present experiments on Lotka-Volterra predator-prey model and Bayesian GAN. 

However, it is not clear how to extend these methods to the high-dimensional setting where choice of summary statistics is crucial. Further, the mean field assumption in \citep{tran2017hierarchical} is not valid for time-series models. In contrast to above methods, this paper clearly demarcates the summarization from the approximation of the non-differentiable simulator. Additionally, not being constrained by a variational setup in \citep{tran2017hierarchical}, one can use sophisticated approximators/summarizer pairs for ``simulating" a black-box simulator.

\paragraph{Maximum Mean Discrepancy}
The Maximum Mean Discrepancy (MMD) is an integral probability metric defined via a kernel $k$ and its associated Reproducing Kernel Hilbert Space (RKHS) \citep{muandet2017kernel,sriperumbudur2012empirical,gretton2007kernel}. Explicitly, the MMD distance with kernel $k$ between distributions $P$ and $Q$ is given by $MMD(k, P,Q) :=  E[k(X,\tilde{X})] - 2 E[k(X,Y)]  + E[k(Y, \tilde{Y})]$ where $X,\tilde{X}$ are independent copies from $P$ and $Y, \tilde{Y}$ are independent copies from $Q$. For empirical samples  $X:=\{x_1,\cdots,x_m\}$ from $P$ and $Y :=\{y_1,\cdots y_n\}$ from $Q$, an unbiased estimate of the MMD is $\widehat{MMD}(k, X, Y) :=  \frac{1}{m(m-1)} \sum_{i,i'} k(x_i,x_{i'}) 
 - \frac{2}{mn} \sum_{i,j} k(x_i,y_j)
+ \frac{1}{n(n-1)} \sum_{j,j'} k(y_j, y_{j'})$.
% \begin{align*} 
% \widehat{MMD}(k, X, Y) := & \frac{1}{m(m-1)} \sum_{i,i'} k(x_i,x_{i'}) \\
% & - \frac{2}{mn} \sum_{i,j} k(x_i,y_j) \\
% & + \frac{1}{n(n-1)} \sum_{j,j'} k(y_j, y_{j'}).
% \end{align*}
As shown in \citet{dziugaite2015training}, this can be differentiated with respect to parameters generating one of the distributions.

% %%% Local Variables:
% %%% mode: latex
% %%% TeX-master: "main"
% %%% End:

\section{Model}
\label{sec:model}

\begin{figure}[tbp]
 \begin{tabular}{cc}
  \fbox{\includegraphics[align=t,width=0.4\linewidth]{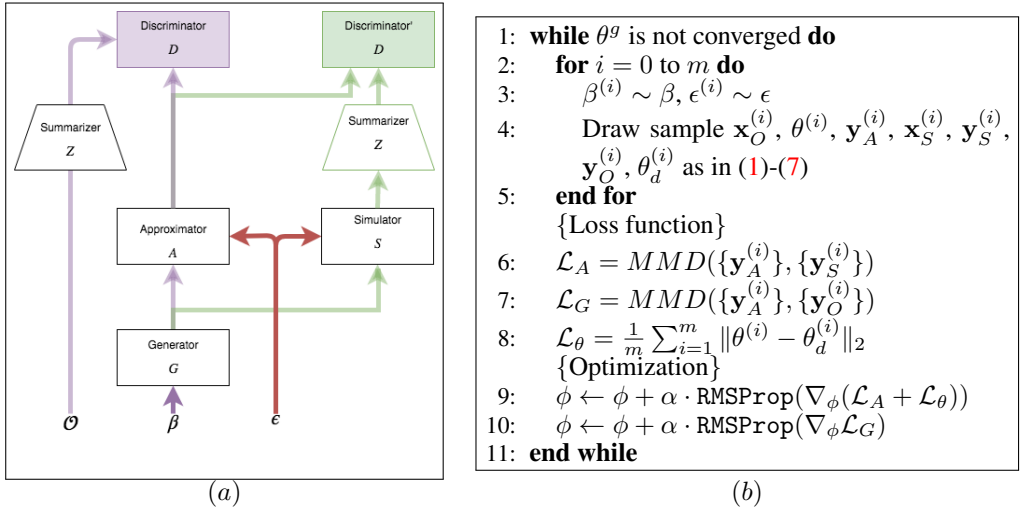}} 
  & 
  \fbox{\begin{minipage}[t]{0.5\linewidth}
     \begin{algorithmic}[1]
    \WHILE{$\theta^{g}$ is not converged}
    \FOR{$i=0$ to $m$}
    \STATE $\beta^{(i)} \sim \beta $, $\epsilon^{(i)} \sim \epsilon $ 
    \STATE Draw sample $\bx_O^{(i)}$, $\theta^{(i)}$, $\by_A^{(i)}$, $\bx_S^{(i)}$, $\by_S^{(i)}$, $\by_O^{(i)}$, $\theta_d^{(i)}$ as in \eqref{eq:obs}-\eqref{eq:decoder}
    \ENDFOR
    
    \COMMENT{Loss function} 
    \STATE ${\cal L}_A= MMD(\{\by^{(i)}_A\}, \{\by^{(i)}_S\})$
    \STATE ${\cal L}_G = MMD(\{\by^{(i)}_A\}, \{\by^{(i)}_O\})$
    \STATE ${\cal L}_\theta = \frac{1}{m} \sum_{i=1}^{m} \| \theta^{(i)} - \theta_d^{(i)} \|_2$
    
    \COMMENT{Optimization} 
    \STATE  $\phi \leftarrow \phi + \alpha  \cdot {\tt RMSProp}(\nabla_\phi ({\cal L}_A + {\cal L}_\theta)) $
    \STATE  $\phi \leftarrow \phi + \alpha  \cdot {\tt RMSProp}(\nabla_\phi {\cal L}_G) $
    \ENDWHILE
    %\STATE Generate fake samples from the physical simulator %$y^{(i)}_{sim}=f(\theta^{(i)}, \epsilon^{(i)})$
    %\STATE Draw samples from the approximator $z^{(i)}=A_v()$
    \end{algorithmic}
  \end{minipage}} \\ 
  $(a)$ & $(b)$ 
 \end{tabular}
  \caption{ $(a)$ ABC-GAN architecture for ABC computation. The external dependence on noise ($\epsilon$) is shown in \textcolor{red}{red}. Two distinct network paths (shown in \textcolor{green}{green} and \textcolor{pink}{pink})  correspond to two different optimizations~(resp., \textcolor{green}{\PA} and \textcolor{pink}{\PB}) akin to ratio test in GANs; $(b)$ ABC-GAN implementation $\alpha=10^{-3}$, $m=50$.}
  \label{fig:normal-gan} \label{alg:abcgan} 
  \end{figure}
 
  % \begin{algorithm}[t]
  %   \caption{ABC-GAN: $\alpha=10^{-3}$, $m=50$}
  %   \begin{algorithmic}[1]
  %   \WHILE{$\theta^{g}$ is not converged}
  %   \FOR{$i=0$ to $m$}
  %   \STATE $\beta^{(i)} \sim \beta $, $\epsilon^{(i)} \sim \epsilon $ 
  %   \STATE Draw sample $\bx_O^{(i)}$, $\theta^{(i)}$, $\by_A^{(i)}$, $\bx_S^{(i)}$, $\by_S^{(i)}$, $\by_O^{(i)}$, $\theta_d^{(i)}$ as in \eqref{eq:obs}-\eqref{eq:decoder}
  %   \ENDFOR
    
  %   \COMMENT{Loss function} 
  %   \STATE ${\cal L}_A= MMD(\{\by^{(i)}_A\}, \{\by^{(i)}_S\})$
  %   \STATE ${\cal L}_G = MMD(\{\by^{(i)}_A\}, \{\by^{(i)}_O\})$
  %   \STATE ${\cal L}_\theta = \frac{1}{m} \sum_{i=1}^{m} \| \theta^{(i)} - \theta_d^{(i)} \|_2$
    
  %   \COMMENT{Optimization} 
  %   \STATE  $\phi \leftarrow \phi + \alpha  \cdot {\tt RMSProp}(\nabla_\phi ({\cal L}_A + {\cal L}_\theta)) $
  %   \STATE  $\phi \leftarrow \phi + \alpha  \cdot {\tt RMSProp}(\nabla_\phi {\cal L}_G) $
  %   \ENDWHILE
  %   %\STATE Generate fake samples from the physical simulator %$y^{(i)}_{sim}=f(\theta^{(i)}, \epsilon^{(i)})$
  %   %\STATE Draw samples from the approximator $z^{(i)}=A_v()$
  %   \end{algorithmic}
  %   \label{alg:abcgan} 
  %   \end{algorithm}

Let $P(\bX | \theta)$ denote a target distribution with unknown parameters $\theta_0\in \Re^d$ that need to be estimated. We have access to a black box simulator~$S:(\theta, \epsilon)\to \bX$ allowing us to generate samples from the distribution for any choice of parameter $\theta$. Here, we have captured the underlying stochasticity of the simulator using suitable noise $\epsilon$.  

Figure~\ref{fig:normal-gan} gives a high-level overview of the ABC-GAN architecture. The inputs to the network are the generator noise $\beta$ and the simulator noise $\epsilon$ which are problem-specific.  The network functions are given as
\begin{align}
\bx_O & \sim {\cal O} &  \textrm{(observations)} \label{eq:obs} \\ 
\theta & = G(\beta | \phi_G) &  \textrm{(generator)}  \\
\by_A & = A(\theta, \epsilon  |  \phi_A)   & \textrm{(approximator)}  \\
\bx_S & = S( \theta, \epsilon ) &  \textrm{(simulator)} \\
\by_S & = Z( \bx_S | \phi_Z) &  \textrm{(summary-sim.)}   \\ 
\by_O & = Z( \bx_O | \phi_Z)&  \textrm{(summary-obs.)}  \\
\theta_d & = A_d(y_A | \phi_{A_d} )&  \textrm{(decoder-approx.)}\label{eq:decoder}
\end{align}  
The parameters are $\phi=\{\phi_G, \phi_A, \phi_Z, \phi_{A_d} \}$ and the optimization consists of two alternating stages, namely: 
\begin{itemize}
\item \PA: This phase trains the approximator $A$ and the summarizer $Z$ against the black-box simulator using a ratio test between $y_A$ and $y_Z$. Mathematically,  
\[ \min_{\phi_A, \phi_Z, \phi_{A_d} } \mathbb{E}_{\beta, \epsilon}  \left[ {\cal L}_A(\by_A, \by_S)   + {\cal L}_{\theta}( \theta, \theta_d) \right] \]
where the loss term ${\cal L}_{\theta}$ corresponds to a decoder for approximator $A$ (as an encoding of parameter $\theta$). 

\item \PB: This phase trains the generator $G$ in order to generate parameters that are similar to the observed data, i.e., with better acceptance rates. Mathematically,
\[ \min_{\phi_G} \mathbb{E}_{\beta, \epsilon, x_O \sim {\cal O}} \left[ {\cal L}_G(\by_A, \by_O)\right] \] 
\end{itemize}

The \PA optimization (without \PB) reduces to function approximation for the summary statistics of black-box simulator, with the decoder loss ensuring the outputs of the approximator and summarizer do not identically go to zero. A network as described above can  be used in place of the simulator in a transparent fashion within an ABC scheme. However, this is extremely wasteful and the \PB scheme incorporates gradient descent in parameter space to quickly reach the acceptance region similar to BOLFI~\citep{gutmann2017likelihood}. Given the perfect approximator (or directly, the black-box simulator) and choosing an $f$-divergence~\citep{nowozin2016f,mohamed2016learning} as loss ${\cal L}_G$ ensures the generator unit attempts to produce parameters within the acceptance region. 

% The discriminator shown in Figure~\ref{fig:normal-gan} reflects the two optimization stages described above with different choices for the loss function (e.g., Wasserstein distance~\citep{arjovsky2017wasserstein}, MMD~\citep{dziugaite2015training}, etc.)~\citep[see discussion in][]{mohamed2016learning}.  

Algorithm~\ref{alg:abcgan} shows our specific loss functions and optimization procedure used in this paper. In our experiments, we observed that training with MMD loss in lieu of the discriminator unit~\citep{dziugaite2015training} yielded the best results~(Section~\ref{sec:experiments}). 
% This is a case of moment-matching distance function~\cite[see][for a discussion]{prangle2017adapting}, and we note that using other distance functions (e.g., $f$-divergences~\cite{nowozin2016f}) is easily possible within our framework.  

\paragraph{Discussion}
\label{sec:discussion}
We emphasize that our implementation (Algorithm~\ref{alg:abcgan}) is just one possible solution in the ABC-GAN architecture landscape. For a new problem of interest, we highlight some of these choices below: 
\begin{itemize}
\item {\em Pretraining (\PA)}:
For a specific region of parameter space~(e.g., based on domain knowledge), one can pre-train and fine-tune the approximator $A$ and summarizer $S$ networks by generating more samples from parameters within the restricted parameter space. 
\item {\em Automatic summarization}: Further, in domains where the summary statistic is not so obvious or rather adhoc~(Section~\ref{sec:ricker}) - \PA optimization provides a natural method for identifying good summary statistics. 
\item {\em Loss function/Discriminator}: \citet{prangle2017adapting} discuss why standard distance functions like Euclidean might not be suitable in many cases. In our architecture, the loss function can be selected based on the insights gained in the GAN community~\citep{DBLP:journals/corr/LiSZ15,DBLP:journals/corr/BellemareDDMLHM17,DBLP:journals/corr/Arora0LMZ17,sutherland2016generative}.
\item {\em Training}: GANs are known to be hard to train. In this work, \PA is more crucial (since the approximator/summarizer pair tracks the simulator) and in general should be prioritized over \PB (which chooses the next parameter setting to explore). We suggest using several rounds of \PA for each round of \PA in case of convergence problems. 

\item {\em Mode collapse}: We did not encounter mode collapse for a simple experiment on univariate normal (Supplementary Material, Section $A$). However, it is a known problem and an active area of research~\citep{arjovsky2017wasserstein,DBLP:journals/corr/Arora0LMZ17} and choosing the Wasserstein distance~\citep{arjovsky2017wasserstein} has yielded promising results in other problems.

\item {\em Module internals}: Our architecture is not constrained by independence between samples (vis-a-vis \citep{tran2017hierarchical}). For example, in time series problems, it makes sense to have deep recurrent architectures (e.g., LSTMs, GRUs) that are known to capture long-range dependencies~(e.g., Sections~\ref{sec:ricker}). Design of network structure is an active area of deep learning research which can be leveraged alongside domain knowledge.
\end{itemize}

\section{Experiments}
\label{sec:experiments} 
All experiments are performed using TensorFlow r1.3 on a Macbook pro 2015 laptop with core i5, 16GB RAM and {\em without} GPU support. The code for the experiments will be  made available on github. In addition to experiments reported below, Section~$A$ in the supplementary material evaluates our ABC-GAN architecture on two small synthethic models, namely, univariate-normal and mixture of normal distributions.

% \url{http://www.bitbucket.org/vjethava/gan/}. 

% \subsection{Synthetic Data}

% \input{normal-exp.tex}

%\input{mixturenormal.tex}

\subsection{Generalized linear model}
\label{sec:glm}
\begin{figure*}[t!]
\includegraphics[width=\linewidth]{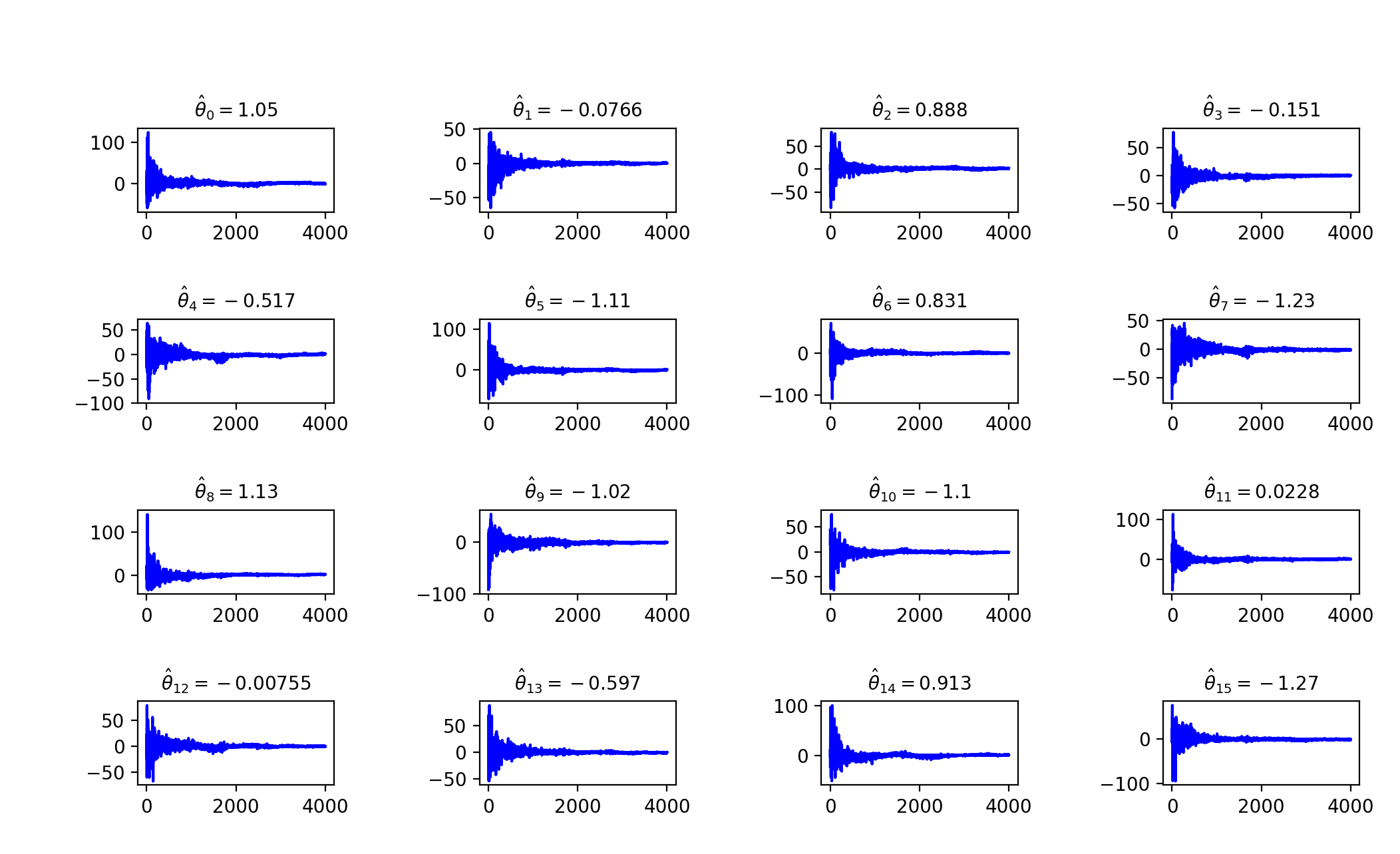}
\caption{Inferred parameter values for one run of ABC-GAN for the GLM model (Section~\ref{sec:glm}) with prior $\theta_i \sim {\tt Uniform}(-100, 100)$ and two layer feed-forward networks used for the generator and approximator in the  ABC-GAN structure. The true parameter value is $\bf 0$. We see that ABC-GAN algorithm recovers $\hat{\theta}$ close to the true parameter $\bf 0$.}
\label{fig:glm-res}
\end{figure*}
\citet{kousathanas2016likelihood} note that basic ABC algorithm and sequential Monte Carlo methods~\citep{sisson2007sequential,beaumont2009adaptive} are useful for low-dimensional models, typically less than $10$ parameters. To tackle higher dimensions, \citet{kousathanas2016likelihood} consider scenarios where it is possible to define sufficient statistics for subsets of the parameters allowing parameter-specific ABC computation. This includes the class of exponential family of probability distributions. 
%However, it is not clear whether such a choice is possible for general models. 

In this example, we consider a generalized linear model defined  by \citet[][Toy model 2]{kousathanas2016likelihood} given as 
$s = {\bf C}\theta + \epsilon$ 
% \begin{equation}
%s = {\bf C}\theta + \epsilon
%\end{equation}
where $\theta\in \Re^n$ denotes the unknown parameter, $\epsilon$ denote multi-variate normal random variable ${\cal N}({\bf 0}, {\bf I}_n)$ and $\bf C$ is a design matrix ${\bf C}  = {\bf B} \cdot {\tt det}( {\bf B}^\top {\bf B})^{-\frac{1}{2n}}$ and 
$${\bf B} =  \begin{bmatrix}
\frac{1}{n} & \frac{2}{n} & \ldots & 1 \\ 
1 & \frac{1}{n} & \ldots & \frac{n-1}{n} \\ 
\vdots & \vdots&  \ddots & \vdots \\ 
\frac{2}{n} & \frac{3}{n} & \ldots & \frac{1}{n} 
\end{bmatrix}.$$
% \begin{align*}
% {\bf C} & = {\bf B} \cdot {\tt det}( {\bf B}^\top {\bf B})^{-\frac{1}{2n}}, \;  \text{where}\\
% {\bf B} &=  \begin{bmatrix}
% \frac{1}{n} & \frac{2}{n} & \ldots & 1 \\ 
% 1 & \frac{1}{n} & \ldots & \frac{n-1}{n} \\ 
% \vdots & \vdots&  \ddots & \vdots \\ 
% \frac{2}{n} & \frac{3}{n} & \ldots & \frac{1}{n} 
% \end{bmatrix}.
% \end{align*}
We use a uniform prior $\theta_i \sim {\tt Uniform}(-100, 100)$ as in
the original work. However, we note that
\citet{kousathanas2016likelihood} ``start the MCMC chains at a normal
deviate ${\cal N}(\theta, 0.01{\bf I})$, i.e., around the true values
of $\theta$.'' The true parameter is chosen as $\theta={\bf 0}$.

We do parameter inference for $n=16$ dimensional Gaussian in the above
setting. Figure~\ref{fig:glm-res} shows the mean of
the posterior samples within each mini-batch as the  the algorithm
progresses~\footnote{For clarity, a larger version of this plot is
  presented in Figure 1 of the supplementary material.}. The total
number of iterations is $4000$ with mini-batch size of $10$ and
learning rate of $10^{-2}$. The algorithm takes $10.24$ seconds. We
reiterate that ABC-GAN does not use model-specific information such as
the knowledge of sufficient statistics for subsets of parameters, and
thus, is more widely applicable than the approach of
\citet{kousathanas2016likelihood}. Concurrently, it also enables
computationally efficient inference in  high-dimensional models -- a
challenge for Sequential Monte Carlo based
methods~\citep{sisson2007sequential,beaumont2009adaptive} and
BOLFI~\citep{gutmann2017likelihood}. 

{\em In the high-dimensional setting, classic ABC methods 
or BOLFI cannot be easily used in contrast to this work.}

% \begin{figure}
% \includegraphics[width=\linewidth]{figures/glm-l1.png}
% \caption{$L1$-loss between the mean value of posterior parameter samples and the true parameter $\theta={\bf 0}$. We see that the $L1$-loss drops close to zero showing that ABC-GAN recovers the parameter without using model specific information.}
% \label{fig:glm-l1}
% \end{figure}

%%% Local Variables:
%%% mode: latex
%%% TeX-master: "main"
%%% End:

\subsection{Ricker's model}
\label{sec:ricker}
\begin{figure}
    \begin{center}
      \begin{tabular}{cc}
      \includegraphics[height=0.5\linewidth]{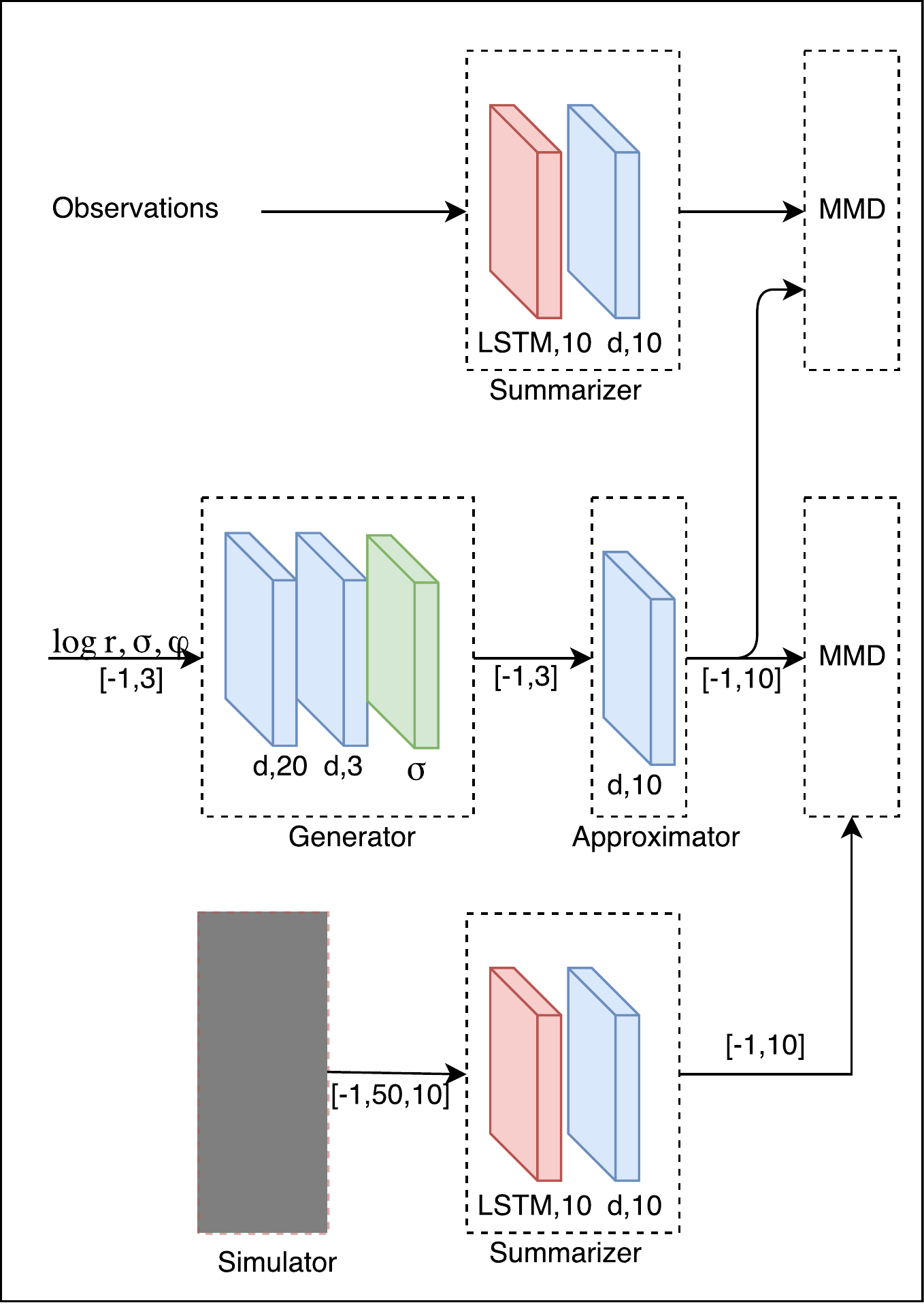} &
      \includegraphics[width=0.5\linewidth,angle=90]{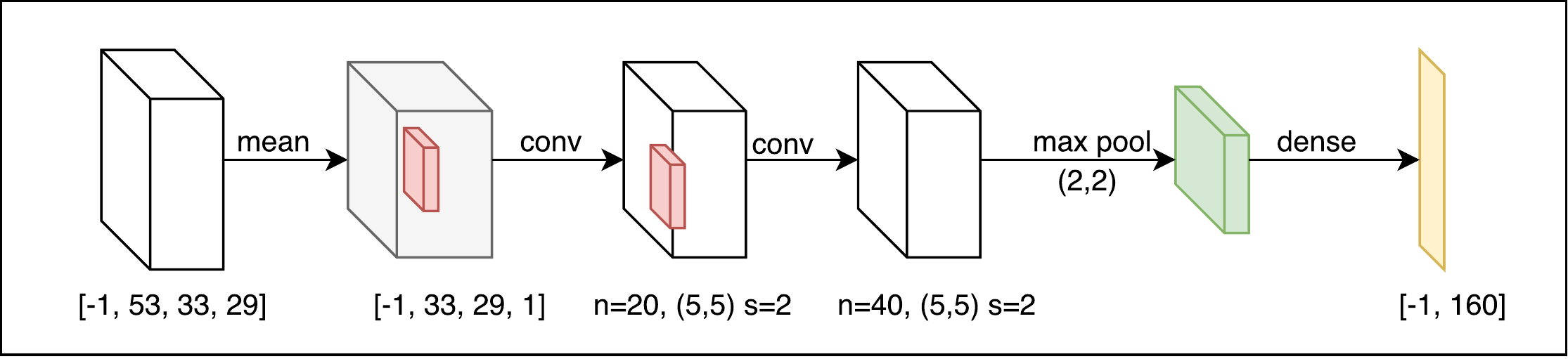}
      \\ $(a)$ Network structure for Ricker model & $(b)$  Convolutional summarizer (DCC)
      \end{tabular}
      \end{center}
    \caption{ $(a)$ Network structure of ABC-GAN for the Ricker model. Here,  $[d,10]$, $\sigma$ and $LSTM,10$  denote  a densely
    connected layer with $10$ outputs, sigmoid activation and a LSTM with
    $10$ units respectively. $(b)$ Convolutional summarizer for the DCC example. We generate summary representation consisting of $160$ features from the input matrices $I^{T_m}_{is}$. The number of filters $n$, size of the filters $(5,5)$ and stride $2$ are shown. The max\_pooling layer has size $(2,2)$. and the final dense layer outputs a summary representation of size $160$ for each input sample.}
      \label{fig:ricker}\label{fig:conv-dcc}    
\end{figure}

The stochastic Ricker
model~\citep{ricker1954stock} is an ecological model described by the nonlinear autoregressive equation: 
$N^{(t)}  =  N^{(t-1)} r \exp \left( - N^{(t-1)}+ \sigma e^{(t)}\right)$ 
%\begin{align}\label{eq:N-t}
%N^{(t)} & =  N^{(t-1)} r \exp \left( - N^{(t-1)}+ \sigma e^{(t)}\right),
%\end{align}
where $N^{(t)}$ is the animal population at time $t\in \{1,\ldots,
n\}$ and $N^{(0)}=0$. The observation $y^{(t)}$ is given by the
distribution $y^{(t)} | N^{(t)}, \phi \sim \texttt{Poisson}(\phi N^{(t)})$,  
%\begin{equation}
%y^{(t)} | N^{(t)}, \phi \sim \texttt{Poisson}(\phi N^{(t)}), 
%\end{equation}
and the model parameters are  $\theta=(\log r, \sigma, \phi)$. The
latent time series $N^{(t)}$ makes the inference of the parameters
difficult. 

\citet{wood2010statistical} computed a synthetic log-likelihood by
defining the following summary statistics: mean, number of zeros in
$y^{(t)}$,  auto-covariance with lag $5$, regression coefficients for
$(y^{(t)})^3$ against $ [ (y^{(t-1)})^3,\, (y^{(t-1)})^6 ]$. They fit a
Gaussian matrix to the summary statistics of the simulated 
samples and then, the synthetic log-likelihood is given by the probability of observed data
under this Gaussian model. \citet{wood2010statistical} inferred the
unknown parameters using standard Markov Chain Monte Carlo~(MCMC)
method based on their synthetic log-likelihood. \cite{gutmann2016bayesian} used the same synthetic
log-likelihood but reduced the number of required samples for
parameter inference based on  regressing 
the discrepancy between simulated and observed samples (in terms of
summary statistics) on the parameters using Gaussian Processes. 

% Our method has the following key differences compared to past
% approaches: 
% \begin{itemize}
% \item We learn summary representation rather than using the adhoc
%   summary statistics defined by \citet{wood2010statistical}.
% \item We use maximum mean discrepancy
%   (MMD), a metric with
%   well-established theoretical properties~\citep{dziugaite2015training,muandet2017kernel} instead of the synthetic
%   likelihood. Other losses and other discriminator networks can be
%   used instead of MMD. 
% \item We explore the parameter space using stochastic gradient descent (\texttt{RMSProp})
%   instead of regression-based approach of
%   \citet{gutmann2016bayesian}. Other optimization algorithms (e.g.,
%   \texttt{Adam}) can be used alternatively.
% \end{itemize}
% Our method allows us to readily leverage the advances in the field of
% deep learning to the task of parameter inference in the ABC setting. 

% \begin{figure}[t!]
% \begin{center}
%   \begin{tabular}{c}
%   \includegraphics[width=0.4\linewidth]{figures/ricker.pdf} \\ 
%   \end{tabular}
%   \end{center}
% \caption{Network structure of ABC-GAN for the Ricker model. Here,  $[d,10]$, $\sigma$ and $LSTM,10$  denote  a densely
% connected layer with $10$ outputs, sigmoid activation and a LSTM with
% $10$ units respectively.}
%   \label{fig:ricker}
% \end{figure}
\begin{figure}
  \centering
\includegraphics[width=0.8\linewidth]{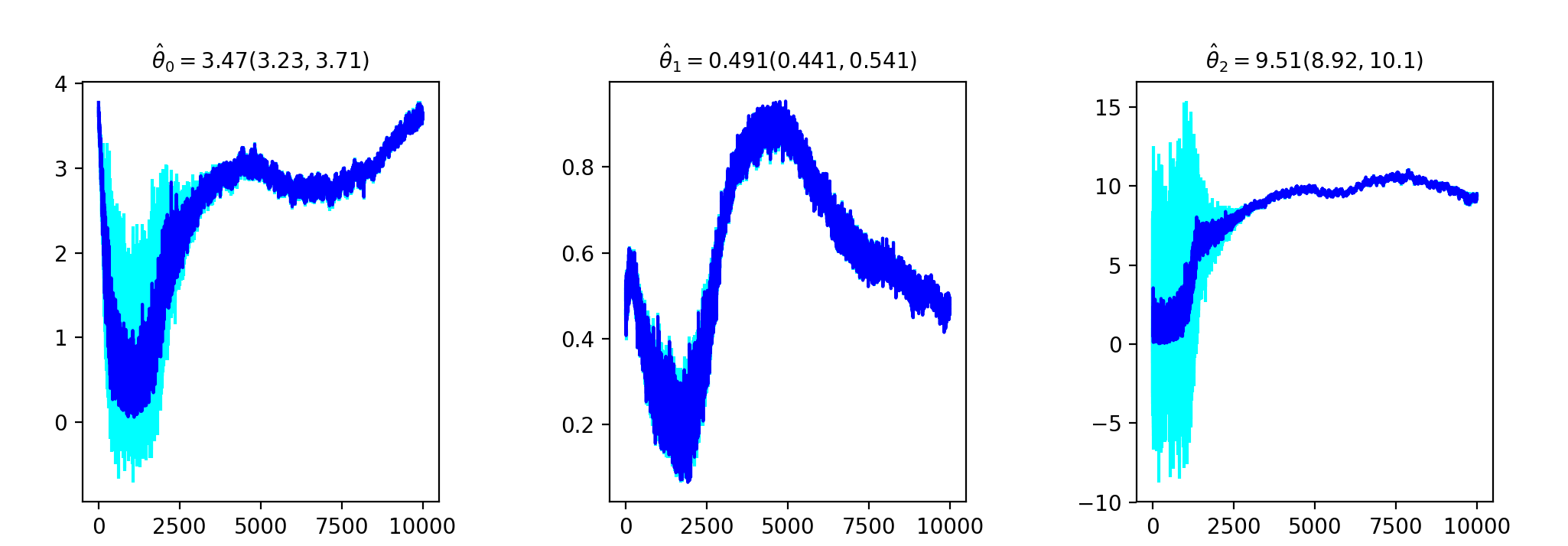} 
\caption{Inferred parameters for the Ricker model~(Section~\ref{sec:ricker}) in a single run after $10000$ iterations using \texttt{RMSProp}
with learning rate=$10^{-3}$, mini-batch size $10$ and sequence length
$10$.  The means of the estimated parameters for this run  are $(3.47, 0.49, 9.51)$ and the $\pm 3\sigma$ range is
shown.}
\label{fig:ricker-parameter}
\end{figure}

\begin{figure}\centering
\includegraphics[width=0.8\linewidth]{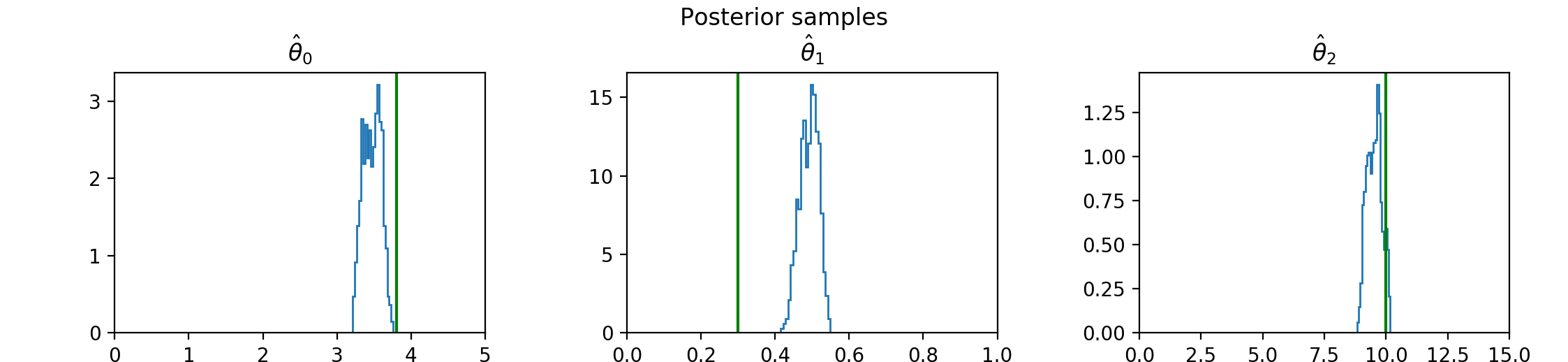}
\caption{The histogram of the posterior samples for the last
  1000 iterations of ABC-GAN in one run for the Ricker model. The true parameters $(3.8, 0.3, 10)$ are shown in \textcolor{green}{green}.}
\label{fig:ricker-posterior}
\end{figure}

Figure~\ref{fig:ricker} shows the network structure for the Ricker model~\footnote{Please see the supplementary material for detailed description of the ABC-GAN architecture for the Ricker model.}
We follow the experimental setup of \citet{wood2010statistical}. We
simulate observations from the Ricker model with true parameters $(3.8,
0.3, 10)$. We use the following prior distributions: 
$$\log r \sim \tt{Uniform}(0, 5), \; \sigma \sim \tt{Uniform}(0,1),\; \phi \sim \tt{Uniform}(0, 15).
$$
% $$\begin{array}{r}
% \log r \sim \tt{Uniform}(0, 5), \\ \sigma \sim \tt{Uniform}(0,1),\\ \phi \sim \tt{Uniform}(0, 15).\end{array} 
% $$
Figure~\ref{fig:ricker-parameter} shows the output of the generator
(posterior samples for the parameters) for $10000$ iterations of the
algorithm. A mini-batch size of $10$ is used with each sequence having
length $10$, and the optimization is done using {\tt RMSProp} with a
learning rate of $10^{-3}$.  Figure~\ref{fig:ricker-posterior} shows the histogram of the posterior samples for the last $1000$ iterations of the algorithm. We note that the algorithm converges and is quite close to the true parameters. 

We obtain the posterior means as $(3.185\pm 0.249,   0.677\pm 0.077,  11.747 \pm 0.767)$ (averaged over $10$ independent runs) for the number
of true observations being $N=50$ and an typical time of $362$ seconds for
$10000$ iterations.  BOLFI estimates the
following parameters  $(4.12, 0.15, 8.65)$ for $N=50$ data
points~\citep[see][Figure 9]{gutmann2016bayesian} and a typical run of
BOLFI takes $300$ seconds for the given setup~\footnote{The code
  provided by Dr. Michael Gutmann uses the GNU R and C code of Wood
  for synthetic log-likelihood and is considerably faster than the
  python version available in ELFI.}.

We re-emphasize that compared to the highly-engineered features of
\citet{wood2010statistical}, our summary representation is learnt
using a standard LSTM-based neural network~\citep{hochreiter1997long,graves2012supervised}. Thus, our
approach allows for easier extensions to other problems compared to
manual feature engineering.

{\em  
For complex simulators with latent variables (in a low-dimensional setting) with less number of observations and limited simulations, BOLFI performs best at significantly higher computational cost. 
Further, \citep{tran2017hierarchical} is not suitable as mean field assumption is violated in the time-series model. 

}
%%% Local Variables:
%%% mode: latex
%%% TeX-master: "main"
%%% End:

\subsection{Infection in Daycare center}
\label{sec:dcc}
\begin{figure}
  \centering
  \includegraphics[width=0.8\linewidth]{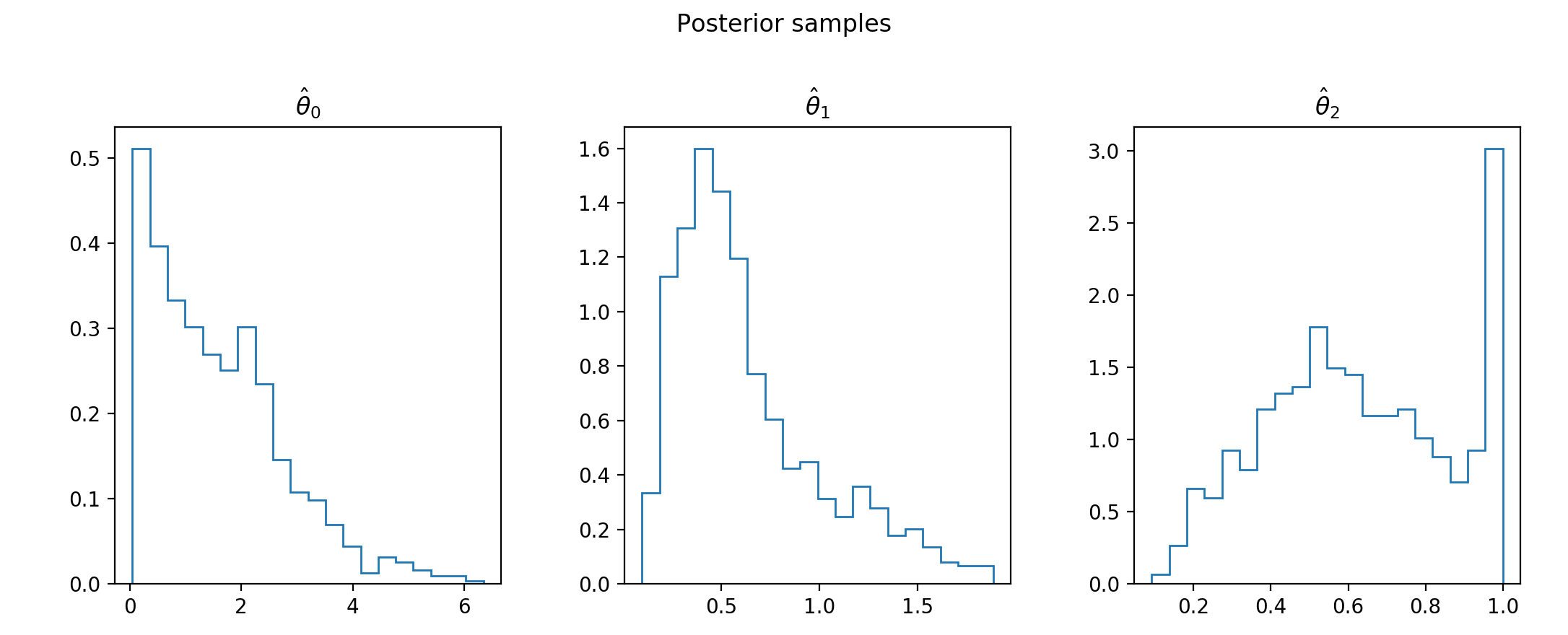}
  \caption{Result for the DCC example using the non-random features of \citet{gutmann2017likelihood}. The
  true parameters are $(3.6, 0.6, 0.1)$. A mini-batch size of $2$ was
  used for $1000$ iterations of the algorithm. The result shows the histogram of the posterior samples generated by the algorithm.}
  \label{fig:daycare1}
\end{figure}
\begin{figure}
  \centering
  \includegraphics[width=0.8\linewidth]{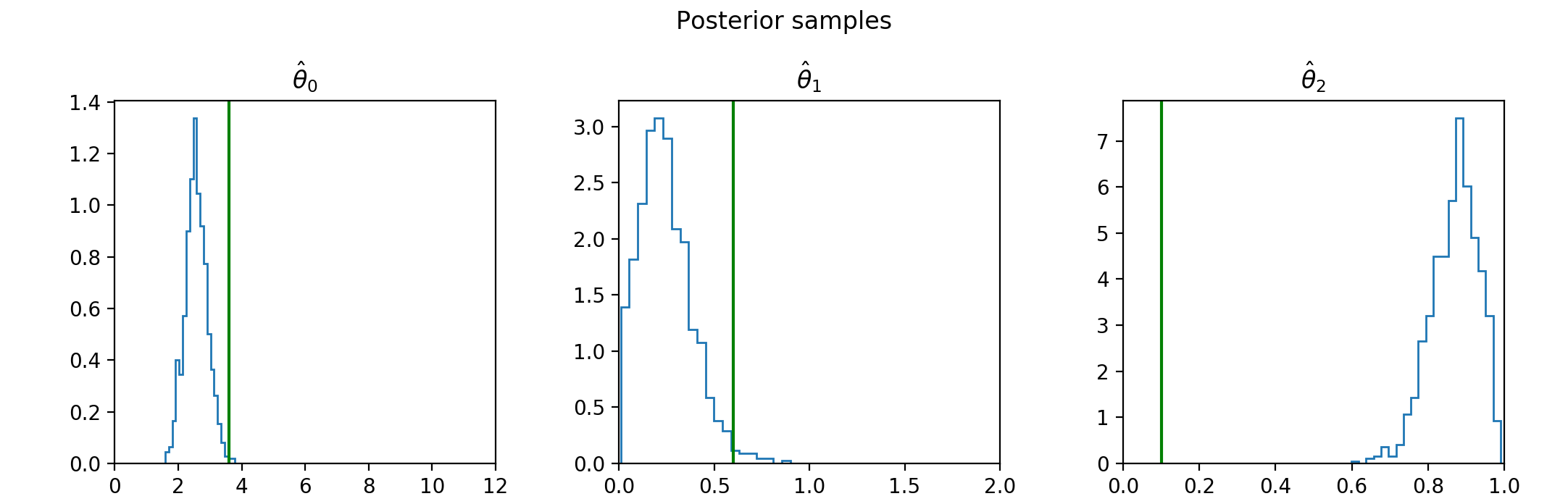}
  \caption{Histogram of the posterior samples generated by ABC-GAN  for the DCC example using the convolutional summarizer. A learning rate of $10^{-3}$ is used for $1000$ iterations.  The true parameters $(3.6, 0.6, 0.1)$ are shown in green.}
  \label{fig:dccConvRes}
\end{figure}
% \begin{figure}[t!]
%   \includegraphics[width=\linewidth]{figures/convSummarizer.pdf}
%   \caption{Convolutional summarizer for the DCC example. We generate summary representation consisting of $160$ features from the input matroces $I^{T_m}_{is}$. The number of filters $n$, size of the filters $(5,5)$ and stride $2$ are shown. The max\_pooling layer has size $(2,2)$. and the final dense layer outputs a summary representation of size $160$ for each input sample.}
%   \label{fig:conv-dcc}
% \end{figure}
We study the transmission of strains of {\em Streptococcus pneumoniae} in a total of $611$ children attending one of $29$ day care centers in Oslo, Norway. The initial data was published by \citet{vestrheim2008phenotypic} and further described in a follow-up study~\citep{vestrheim2010impact}.
\citet{numminen2013estimating} first presented an ABC-based approach
for inferring the parameters associated with rates of infection from
an outside source~($\Lambda$), infection from within the DCC~($\beta$)
and infection by multiple strains
($\theta$). \citet{gutmann2017likelihood} presented a
classification-based approach where they used classification as a
surrogate for the rejection test of standard ABC method. Section~C in
the supplementary material presents a discussion of the
hand-engineered features of \citet{numminen2013estimating} and
\citet{gutmann2017likelihood}. 

In the remainder of this section, we follow the nomenclature in \citep{gutmann2017likelihood}. For a single DCC, the observed data consists of presence or absence of a particular strain of the disease at time $T^m$ when the swabs were taken. On average, $N=53$ individuals attend a DCC out of which only some are sampled. There are $S=33$ strains of the bacteria in total. So the data from each of the $M=29$ DCCs consists of a binary matrix with entries, $I^t_{is}$ where $I^t_{is}=1$ if attendee $i$ has strain $s$ at time $t$ and zero otherwise. The observed data $\bf X$ consists of a set of $M=29$ binary matrices formed by $I^{T_m}_{is}, i=1,\ldots, N_m, s=1,\ldots, 33$. For the simulator, we use the code provided by Michael Gutmann which in turn uses the code of Elina Numminen~\footnote{\url{https://www.cs.helsinki.fi/u/gutmann/code/BOLFI/}}. 

We assume the following prior on the parameters: 
$$ \Lambda \sim {\tt Uniform}(0, 12),\;  \beta \sim {\tt Uniform}(0, 2),\;  \theta \sim {\tt Uniform}(0, 1). $$
% $$\begin{array}{c}
% \Lambda \sim {\tt Uniform}(0, 12), \\
% \beta \sim {\tt Uniform}(0, 2),\\ 
% \theta \sim {\tt Uniform}(0, 1). 
% \end{array}
% $$
%
% \begin{figure}
%   \centering
%   \includegraphics[width=\linewidth]{figures/lr1e-2_dccGutmann200.png}
%   \caption{Result for the DCC example. The
%   true parameters are $(3.6, 0.6, 0.1)$. A mini-batch size of $2$ was
%   used for $200$ iterations of the algorithm.}
%   \label{fig:daycare1}
% \end{figure}
%
We first use the non-random features defined by \citet{gutmann2017likelihood} as our summary statistics. Figure~\ref{fig:daycare1} shows the histogram of the  posterior samples generated by ABC-GAN using the non-random features defined by \citet{gutmann2017likelihood}. We note that the results are not encouraging in this case, though this is an artifact of our algorithm. In order to address this, we define a new summary representation using a convolutional network. 
%Figure~\ref{fig:daycare1} shows the progress of the algorithm for $200$ iterations where the true parameters are $(3.6, 0.6, 0.1)$. A learning rate of $10^{-2}$ and mini-batch size of $2$ is used in the example. 
Figure~\ref{fig:conv-dcc} shows the structure of the convolutional
summarizer which is used to generate summary representations instead
of the non-random features defined by
\citet{gutmann2017likelihood}. The generator and approximator modules
are standard one-layer feed forward networks which are fully described in
Section~C of the supplementary material.

Figure~\ref{fig:dccConvRes} shows the results for the convolutional summarizer. We note that the posterior samples improve especially for the parameters $\Lambda=3.6$ and $\beta=0.6$. However, there is considerable room for improvement by using alternative summarization networks and improved training especially using more rounds of \PA (for better approximation). We leave this to future work.  

  %%% Local Variables:
%%% mode: latex
%%% TeX-master: "main"
%%% End:

\section{Conclusions}
\label{sec:conclusions}
We present a generic architecture for training a differentiable approximator module which can be used in lieu of black-box simulator models without any need to reimplement the simulator. Our approach allows automatic discovery of summary statistics and crystallizes the choice of distance functions within the GAN framework. The goal of this paper is {\em not} to perform "better" than all existing ABC methods under all settings, rather to provide an easy recipe for designing scalable likelihood-free inference models using popular deep learning tools. 

%%% Local Variables:
%%% mode: latex
%%% TeX-master: "main"
%%% End:

\bibliographystyle{plainnat}
\bibliography{gan.bib}

\end{document}

\maketitle

\tableofcontents

\appendix
\section{Small-scale synthetic experiments}

\subsection{Univariate normal distribution}
\label{sec:univariate-normal}
We generate $N=1000$ observations from the univariate normal distribution with mean $\mu_0=3$ and variance $\sigma^2_0=1$. We use the following priors: 
\begin{equation}
\mu \sim Unif(0, 5), \qquad \sigma^2 \sim Unif(1, 5), 
\end{equation}
and perform parameter estimation ($\hat{\mu}$ and $\hat{\sigma}^2)$ using rejection sampling, BOLFI and ABC-GAN. Table~\ref{tab:univariate-normal-example} shows the timing (in seconds) and accuracy results for the three approaches. We note that BOLFI is a complex method and is unsuited for this case since generation of univariate normal samples is computationally very cheap. Nonetheless, we observe that ABC-GAN converges in reasonable time and gives comparable results to others approaches. 
\begin{table}
\begin{center}
% \begin{tabular}{c}
% \includegraphics[width=\linewidth,valign=c] 
% {figures/musigma_gan_example.eps} \\  
\begin{tabular}{|c|c|c|c|}
\hline
Method& $\hat{\mu}$ & $\hat{\sigma}^2$ & $t$ (in seconds)\\\hline
ABC & 2.92 & 1.17 & 0.147 \\
BOLFI& 2.68 & 1.13 & 	2700  \\
GAN & 3.023 & 1.04 & 37.12 \\\hline
% \end{tabular} \\ 
\end{tabular}
\end{center}
\caption{We compare the results with naive ABC
  (using rejection sampling for $10000$ samples) and BOLFI
  algorithm~(Rejection sampling and ABC were implemented using
  ELFI~\citep{kangasraasio2016elfi}).} 
% \caption{(a) Example of ABC-based learning of mean $\mu$ (shown in
%   \textcolor{blue}{blue}) and variance $\sigma^2$ (shown in
%   \textcolor{red}{red}) of univariate normal using generative
%   adversarial network (GAN). (b) We compare the results with naive ABC
%   (using rejection sampling for $10000$ samples) and BOLFI
%   algorithm~(Rejection sampling and ABC were implemented using
%   ELFI~\citep{kangasraasio2016elfi}).} 
\label{tab:univariate-normal-example}
\end{table}

{\em 
For low-dimensional problems with i.i.d.~samples where simulation cost is cheap, classic ABC methods are best 
in terms of ease of implementation and wall-clock time.
}

\subsection{Mixture of normals}
\label{sec:mixturenormal}
We consider a mixture of two normal distributions, first studied in \citet{sisson2007sequential}, with the observations generated from: 
\begin{equation}
P_\theta(\theta)\sim \frac{1}{2}{\cal N}(0, \frac{1}{100}) + \frac{1}{2}{\cal N}(0, 1)
\end{equation}
Here the second term implies large regions of low probability in comparison to the first term. Both classic rejection sampling and Monte Carlo ABC suffer from low acceptance rate and consequently, longer simulation runs (respectively, $400806$ and $75895$ steps) when generating samples from the low  probability tail region. 

In contrast, we run ABC-GAN for $5000$ iterations with a mini-batch
size of $10$ and sequence length $10$, using a learning rate
$10^{-3}$.

\begin{figure*}[t!]
\begin{center}
\begin{tabular}{cc}
\includegraphics[width=0.5\linewidth]{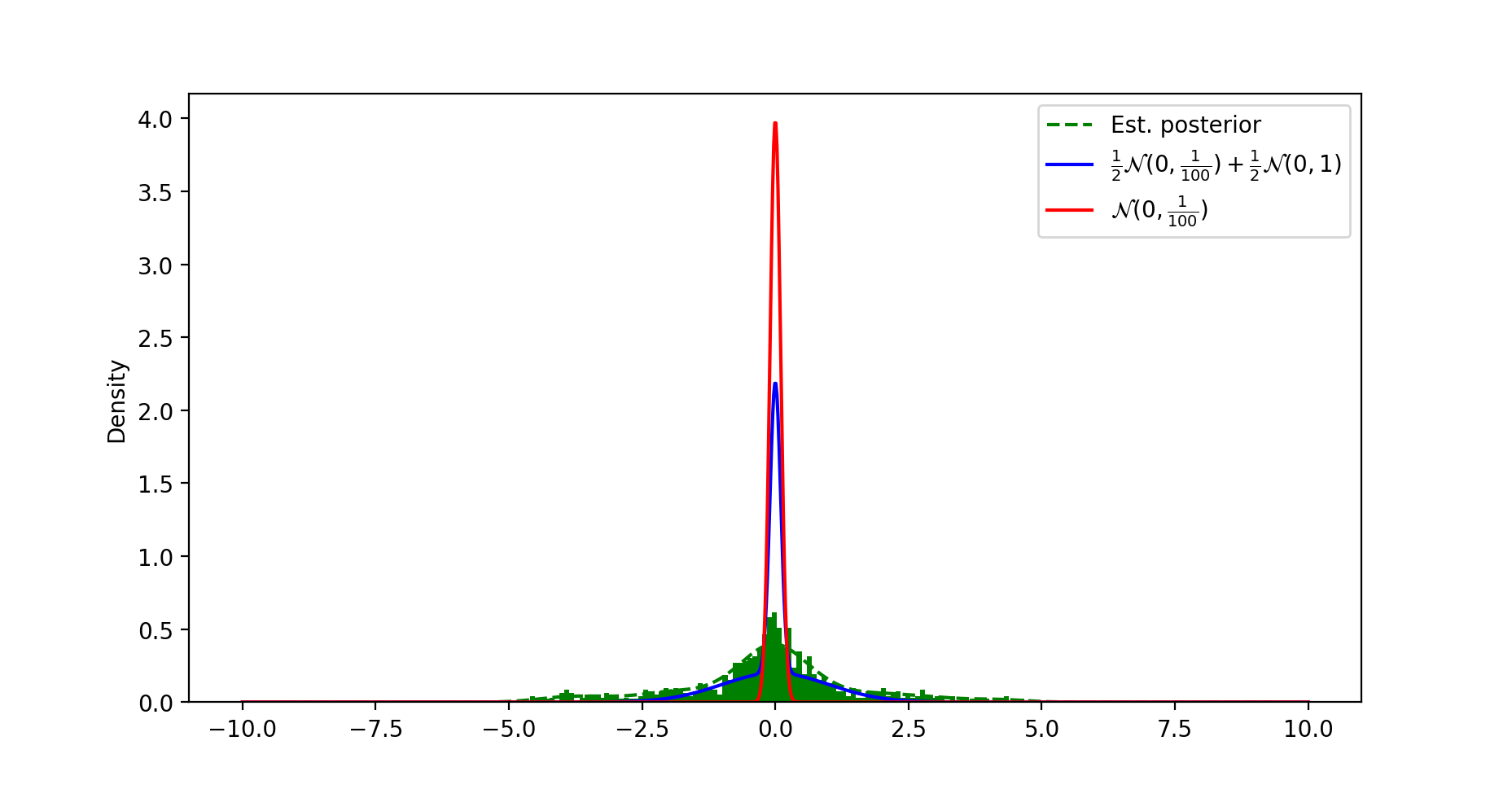} & \includegraphics[width=0.35\linewidth]{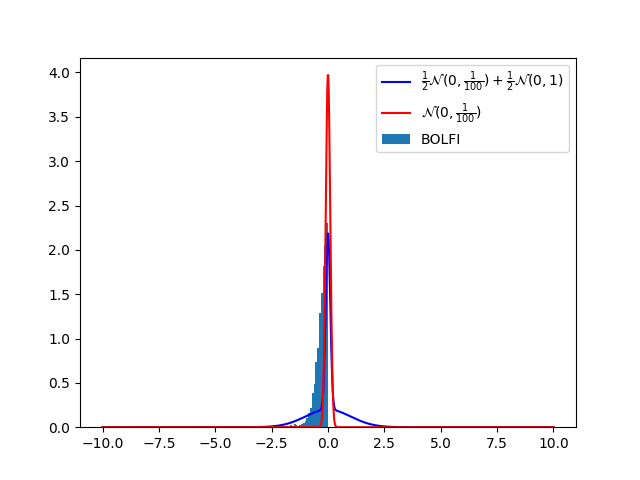} \\ 
(a) & (b) \\
\end{tabular}
\end{center}
\caption{$(a)$ Posterior samples for ABC-GAN (shown in \textcolor{green}{green}) for one run  in the mixture of normals experiment (Section~\ref{sec:mixturenormal}) with $5000$ iterations with a mini-batch size of $10$ and sequence length $10$, using a learning rate $10^{-3}$. The probability density function for the mixture normal and the low-variability component are shown in \textcolor{red}{red} and \textcolor{blue}{blue} respectively. $(b)$ Histogram of posterior samples for first run of BOLFI. Total number of samples is $5000$. We note that the low-probability space is not captured by BOLFI compared to ABC-GAN.}
\label{fig:mixnormal}

\end{figure*}

Figure~\ref{fig:mixnormal} shows the generated posterior samples and
we see that the posterior samples capture the low probability
component of the probability density function. We also run BOLFI~\citep{kangasraasio2016elfi} in order to compare its
result to ABC-GAN.

Table~\ref{tab:mvn} presents the timing and accuracy for the different
methods. The accuracy is specified in terms of
Kullback-Liebler divergence~\citep{cover2012elements} between the true pdf and the empirical pdf
estimated using posterior samples in the range [-10, 10]. We note that the KL-divergence of BOLFI w.r.t. true pdf is lower compared to the KL divergence of ABC-GAN w.r.t. true pdf. However, this does not translate to capturing the low-probability space as shown in Figure~\ref{fig:mixnormal}~(b) where BOLFI does not have samples in the low-dimensional space.

We note that BOLFI fails to adequately capture the
low-probability space (due to the second component) of the
problem. This is reflected in Table~\ref{tab:mvn}. Additionally, the
BOLFI implementation in ELFI is considerably slower than ABC-GAN and
this scenario is exacerbated as more number of observations are
provided or more simulation samples are generated due to the use of
Gaussian Processes. 

% \begin{figure}[t!]
% \includegraphics[width=\linewidth]{figures/mixture_normal.png}
% \caption{Posterior samples for ABC-GAN (shown in \textcolor{green}{green}) for one run (Section~\ref{sec:mixturenormal}) with $5000$ iterations with a mini-batch size of $10$ and sequence length $10$, using a learning rate $10^{-3}$. The probability density function for the mixture normal and the low-variability component are shown in \textcolor{red}{red} and \textcolor{blue}{blue} respectively.}
% \label{fig:mixnormal}
% \end{figure}
% \begin{figure}[t!]
% \includegraphics[width=\linewidth]{figures/mixturenormal_n5000_elfi.png}
% \caption{Histogram of posterior samples for first run of BOLFI for mixture of normals examples (Section~\ref{sec:mixturenormal}). Total number of samples is $5000$. We note that the low-probability space is not captured by BOLFI compared to Figure~\ref{fig:mixnormal}.  }
% \end{figure}

\begin{table}[t!]
\begin{center}
  \begin{tabular}{cccc}\hline
    Method &   $n_{s}$ & $D_{KL}(p\|q)$& $t$ \\\hline
    rej. sampl.$^\dagger$ &  5000 & $0.567\pm 0.012$ & 0.22 \\ 
    ABC-GAN &  5000 & $0.539 \pm 0.196 $ & 14.70 \\ 
    BOLFI$^\dagger$  & 1000 & $0.627 \pm 0.417$ & 126.71 \\
    BOLFI$^\dagger$  & 5000 & $0.445\pm 0.027$ & 426.03 \\ \hline
  \end{tabular}
  \end{center}
  \caption{Timing and accuracy information for the mixture of normals
    example averaged over $10$ independent runs. The  the number of
    samples generated by the simulator ($n_{s}$) are shown. The
    Kullback-Liebler divergence  ($D_{KL}(p\| q)$) between the true pdf $(p)$ and the empirical
  pdf estimated using accepted samples ($q$) and the time taken (for the first run)
in seconds ($t$) are shown. $^\dagger$ELFI returns only negative samples so we add $-x$ to the samples for each original sample $x$ before computing the KL divergence.}
  \label{tab:mvn}
\end{table}

\subsection{Inference in multi-variate normal}
\label{sec:mvn}
We consider multivariate normal model ${\bf X} \in \Re^{16}$ with true mean $\bf 1$ and prior on mean $x_i \sim {\tt Uniform}(0, 10)$. We use two layer feed-forward neural networks as our generator and approximator units.  A learning rate of $10^{-3}$ and a hidden layer of size $16$ is used in the approximator and generator units.
MMD Loss is used as a surrogate for the discriminator as well as for comparing the output of the approximator unit with the simulator. 

\begin{figure}
\centering
\includegraphics[width=\linewidth]{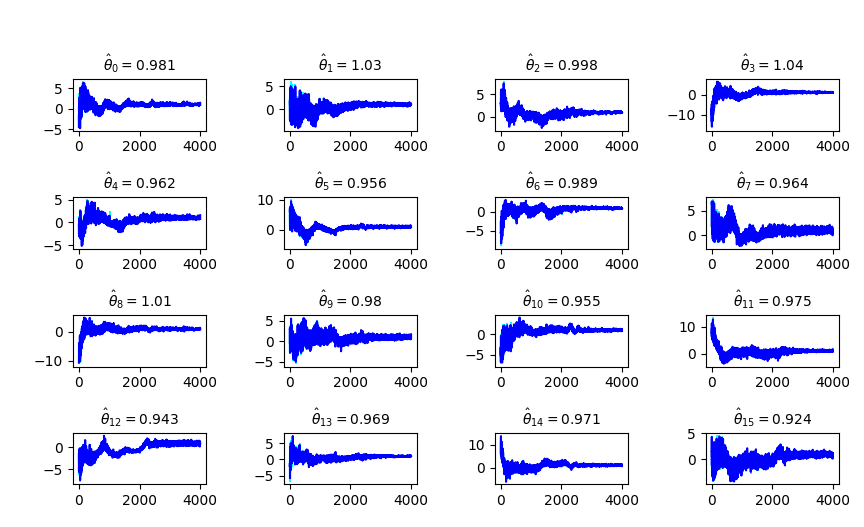}
\caption{Inferred parameters for one run of ABC-GAN for the multi-variate normal model~(Section~\ref{sec:mvn}). We see that the inferred parameters are close to the true mean $\bf 1$.}
\label{fig:mvn_params}
\end{figure}

\begin{figure}
\centering
\includegraphics[width=0.75\linewidth]{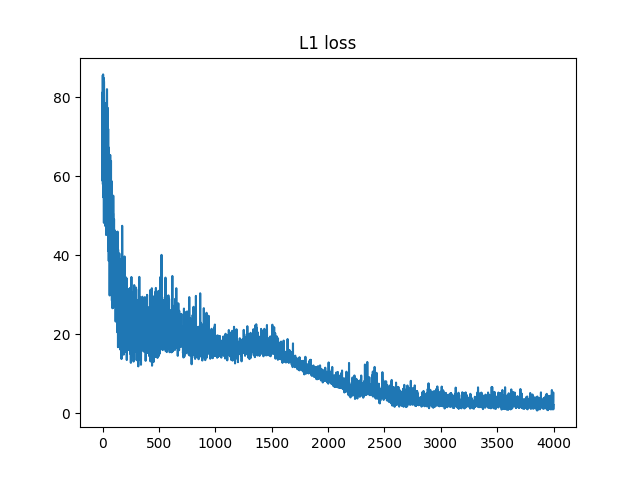}
\caption{The L1-loss between the inferred and the true mean in the multi-variate normal example (Section~\ref{sec:mvn}) as the algorithm progresses.}
\label{fig:mvn_loss}
\end{figure}

Figures~\ref{fig:mvn_params} and \ref{fig:mvn_loss} show the inferred parameters and the L1-loss between the inferred mean and the true mean as the algorithm progresses. We note that the inferred parameters are close to the true mean $\bf 1$. The total time taken is 46.38~seconds.

\section{ABC-GAN specification for Ricker's model}
This section presents the complete model specification of our GAN architecture for the Ricker's model. We describe each of the individual sub-networks below. 

For sake of completeness, we recall the prior distribution on the parameters $\theta=(\log r, \sigma, \phi)$ given by: 
\begin{equation*}
\begin{array}{r}
\log r \sim \tt{Uniform}(0, 5), \\ \sigma \sim \tt{Uniform}(0,1),\\ \phi \sim \tt{Uniform}(0, 15).
\end{array} 
\end{equation*}
In the remainder of this section, the notation $(d=10)$ indicates the
layer has $10$-dimensional output. 
\paragraph{Generator}
The generator $G_u(\cdot)$ takes as input the samples from the prior distribution on $\theta=(\log r, \sigma, \phi)$. The generator has the following equations: 
\begin{align*}
  x_{G,1} & = {\tt ReLu}(w_{g,1}^\top \theta + b_{g,1}) & (d=20)\\
  x_{G,2} &= w_{u,2}^\top x_{g, 2} + b_{g,2} & (d=3) \\
  x_{G,3} &= 5 \times {\tt Sigmoid}(x_{G, 2}[:, 1]) & (d=1) & &(\hat{\log r})\\
  x_{G,4} &=  {\tt Sigmoid}(x_{G, 2}[:, 2]) & (d=1) & & (\hat{\sigma})\\
  x_{G,5} &= 15 \times {\tt Sigmoid}(x_{G, 2}[:, 3]) & (d=1) & & (\hat{\phi} )\\
  x_G &= [x_{G,3}, \, x_{G, 4}, \, x_{G, 5}]  & (d=3) & & (\hat{\theta}_{post})
\end{align*}
The complete set of weights are given by $W_G=\{w_{g,i}, b_{g,i} \forall i\in
1,2\}$ which are initialized using samples from the normal
distribution  ${\cal N}(0, 1)$. 

\paragraph{Approximator}
The approximator unit consists of a feed-forward layer which takes as
input the posterior parameters $x_G$ from the generator and outputs
approximate statistics given as:
\begin{align*}
x_A &= w_{a,1}^\top x_G + b_{a,1} & (d=3)
\end{align*}
where the weights $W_A=\{ w_{a,1} , b_{a,1}\}$ are initialized from
the normal distribution ${\cal N}(0, 1)$. 

\paragraph{Summarizer}
The summarizer unit consists of LSTM cell followed by a dense layer
which projects the sequence to the summary representation. We recall
that the standard LSTM unit is defined as:
\begin{align*}
  f_t &= \sigma (W_f \cdot [h_{t-1}, x_t] + b_f)  & \text{(forget gate)}\\
  i_t &= \sigma ( W_i \cdot [h_{t-1}, x_t] + b_i ) & \text{(input gate)}\\
  \tilde{C}_t &= tanh(W_c \cdot [h_{t-1}, x_t] + b_C) \\
  C_t& = f_t * C_{t-1} + i_t * \tilde{C}_t & \text{(cell state)}\\
  o_t & = \sigma ( W_o \cdot [h_{t-1}, x_t] + b_o) & \text{(output)} \\
  h_t & = o_t * tanh (C_t)  
\end{align*}
where $W_{LSTM}=\{W_f, W_i, W_C, W_o , b_f, b_i, b_C, b_o \}$ are the
weights for a single unit. We use the notation ${\tt LSTM}_k$ to denote
$k$ LSTM units connected serially. 

Given input sequence $seq$ obtained from the simulator or from
observed data, the summarizer output is given by: 
\begin{align*}
  x_{S,1}&= {\tt LSTM}_k(seq) & (d=k=10) \\
  x_S & = w_{s,1}^\top x_{S,1} + b_{s,1} & (d=10)
\end{align*} 
where the weights are given by $W_S=\{w_{S,1} , b_{S,1}\}\cup
w_{LSTM}$. The weights are initialized to
normal distribution and the initial cell state is set to zero tuple of
appropriate size.

\paragraph{Optimization} 
The connections are described as follows:
\begin{align*} 
  x_G &= {\cal G}( \theta; W_G) \\
  x_A &= {\cal A}(x_G; W_A) \\
  {\tt seq}_S &= {\tt simulator}(x_G, \epsilon) \\
  x_S &= {\cal S}({\tt seq}_S ; W_S)\\ 
  x_O &= {\cal S}({\tt seq}_O; W_S )
\end{align*}
where ${\tt seq}_O$ denotes the observed data, $x_O$ is its summary
representation obtained using the summarizer, and $\epsilon$ denotes
normal noise for input to the simulator alongwith the posterior
parameter $X_G$. 

We use maximum mean discrepancy~(MMD)
loss~\citep{dziugaite2015training} to train the network. Given
reproducing kernel Hilbert space~(RKHS)~${\cal H}$ with associated
kernel $k(\cdot, \cdot)$ and data $X, X', X_1,
\ldots, X_N$ and $Y, Y', Y_1, \ldots, Y_M$ from distributions $p$ and $q$
respectively, the maximum mean discrepancy loss is given by:
\begin{align*}
  MMD^2[{\cal H}, p, q] &= E[ k(X, X') - 2k(X, Y) + k(Y, Y')].
\end{align*}
We use the Gaussian kernel and define the following loss functions: 
\begin{align*}
  {\cal L}_G = MMD( x_A, x_O) \\
  {\cal L}_R = MMD( x_A, x_S) 
\end{align*} 
Let $W_R=(W_S, W_A)$ denote the combined weights for the summarizer
and the approximator units. The update equations for the network are
then given by the following alternating optimizations:
\begin{align*}
  W_R^{(i+1)} &\leftarrow W_R^{(i)} + \alpha \cdot {\tt RMSProp}( \nabla_{W_A, W_S} {\cal L}_R) \\
  W_G^{(i+1)} &\leftarrow W_G^{(i)} + \alpha  \cdot {\tt RMSProp}( \nabla_{W_G}{\cal L}_R ) \\
\end{align*}
where $\alpha$ is the learning rate for the RMSProp algorithm.

\section{Hand-engineered features for the DCC model}

\citet{numminen2013estimating} defined four summary statistics per day care center in order to characterize the simulated and observed data and perform ABC-based inference of the parameters ${\Theta}=(\Lambda, \beta, \theta)$:
\begin{itemize}
\item The strain diversity in the day care centers, 
\item The number of different strains circulating, 
\item The fraction of the infected children, 
\item The fraction of children infected with multiple strains.
\end{itemize}

 \citet{gutmann2017likelihood} classified the simulated data as being fake (not having the same distribution as the observed data) or based on the following features: 
\begin{enumerate}
\item $L2$-norm of the singular values and the rank of the original matrix (2 features),
\item The authors computed the fraction of ones in the set of rows and columns of the matrix. Then, the average and the variability of this fraction was taken across the whole set of rows and columns. Since the average is the same for the set of rows and columns, this yields 3 features.
\item The same features as $(2)$ above for a randomly chosen sub-matrix having $10\%$ of the elements of the original matrix (2 features). 
\end{enumerate}
 By choosing $1000$ random subsets, the authors converted the set of
 $29$ matrices to a set of $1000$ seven-dimensional features. This
 feature set was used to perform classification using LDA. They also did the classification without the randomly chosen subsets mapping each simulated dataset to a five-dimensional feature vector.

\section{ABC-GAN specification for the DCC model}
This section presents the complete description for our architecture doing likelihood free inference in the Daycare center example. We describe each of the sub-networks below.
\paragraph{Generator}
The generator ${\cal G}(\cdot)$ takes as input the samples from the
prior distribution on $\Theta=(\Lambda, \beta, \theta)$. It is
specified as follows:
\begin{align*}
  x_{G,1} & = w_{g,1}^\top \Theta + b_{g,1} & (d=3) \\
  \hat{\Lambda} &= 12 \cdot \sigma( x_{G,1}[:, 0]) & (d=1) \\
  \hat{\beta} &= 2 \cdot \sigma( x_{G,1}[:, 1]) & (d=1) \\
  \hat{\theta} &=  \sigma( x_{G,1}[:, 2]) & (d=1) \\
  x_G &= [ \hat{\Lambda} \; \hat{\beta}\;  \hat{\theta}] 
\end{align*}
The set of variables are given by $W_G=[ w_{g,1}, b_{g,1}]$ which are
initialized using normal distribution.

\paragraph{Approximator}
The approximator unit has a feed-forward network given by:
\begin{align*}
x_A &= w_{a,1}^\top x_G + b_{a,1} 
\end{align*}
where $W_A=[w_{a,1}, b_{a,1}]$ are the weights which are initialized
using the normal distribution.

\section{Supplemental Figures}
\begin{figure*}
\includegraphics[width=\linewidth]{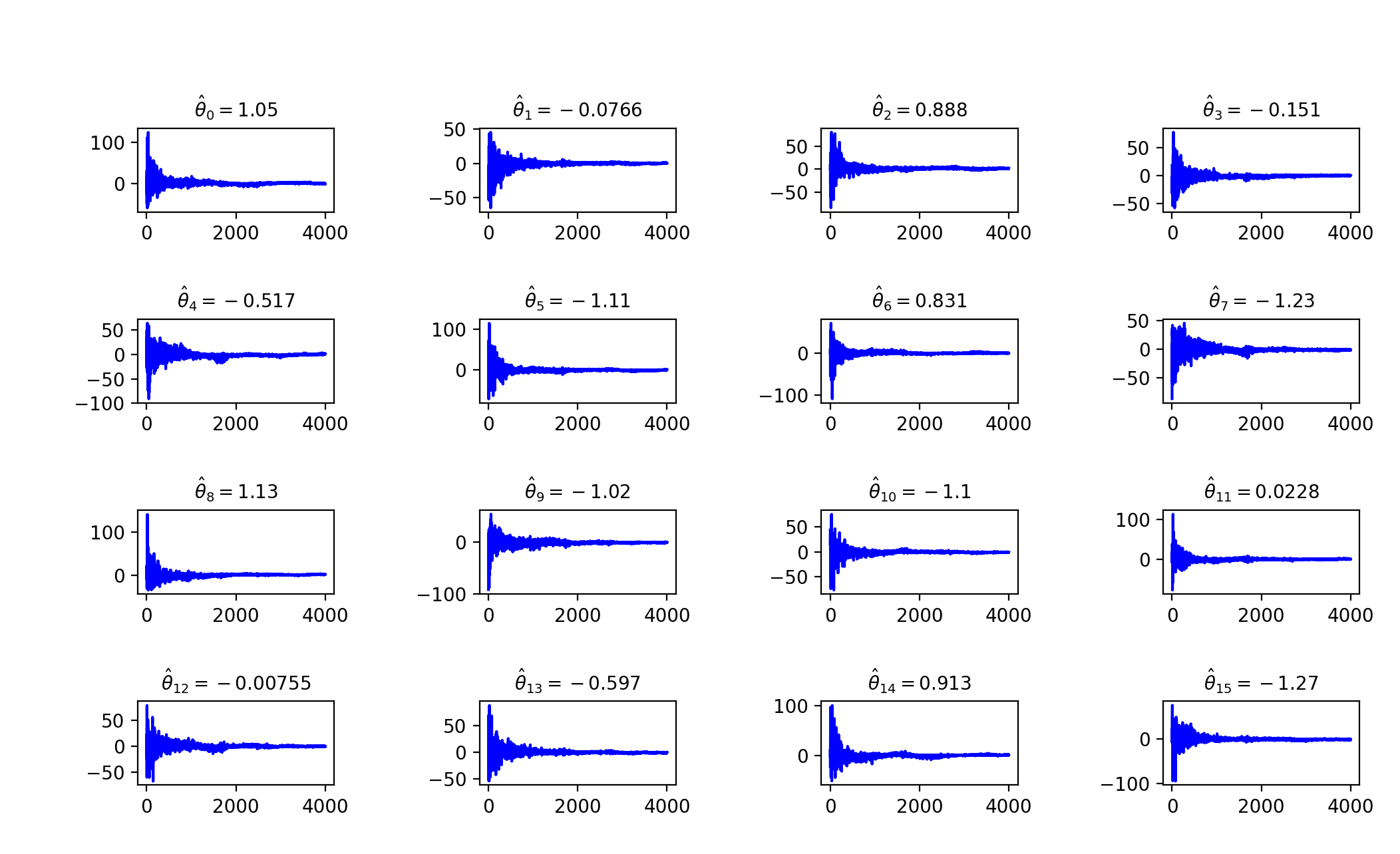}
\caption[]{Inferred parameter values (true parameter is $\bf 0$) with prior $\theta_i \sim {\tt Uniform}(-100, 100)$ and two layer feed-forward networks used for the generator and approximator in the  ABC-GAN structure.}
\label{fig:glm-res}
\end{figure*}

\begin{itemize}
\item Figure~\ref{fig:glm-res} shows the  inferred parameter values in
  Section~4.1.3 of the  main manuscript. The true parameter is $\bf 0$ with
  prior $\theta_i \sim {\tt Uniform}(-100, 100)$ and two layer
  feed-forward networks used for the generator and approximator in the
  ABC-GAN structure.

%   \item Figure~\ref{fig:bolfi-mvn} shows the histogram of the posterior samples generated by BOLFI for the mixture of
%   normals experiment in Section~4.1.2 of the main manuscript.  We note
%   that BOLFI fails to capture the low-probability space adequately. 
\end{itemize}

% \begin{figure*}
% \begin{tabular}{c}
%   \includegraphics[width=0.8\linewidth]{figures/bolfi-mvn-100.png} \\ (a) $n_{o}=100, n_{s}=1000, D(p\|q)=1.336$ \\
%   \includegraphics[width=0.8\linewidth]{figures/bolfi-mvn-100-5000.png} \\ (b) $n_{o}=100, n_{s}=5000, D(p\| q)=1.426 $ \\
%       \includegraphics[width=0.8\linewidth]{figures/bolfi-mvn-1000.png} \\ (c) $n_{o}=1000, n_{s}=1000, D(p\| q)=1.478$ \\
%   \end{tabular}
% \caption[]{Posterior samples generated by BOLFI for the mixture of
%   normal experiment. We see that as the number of samples increases,
%   BOLFI result focusses more on the low-variability component. The
%   number of observations ($n_{o}$), number of samples ($n_{s}$) and
%   KL-divergence between true p (shown in \textcolor{blue}{blue}) and
%   empirical pdf is shown.}
% \label{fig:bolfi-mvn}
% \end{figure*}

\bibliographystyle{plainnat} 
\bibliography{gan.bib}